\documentclass[letterpaper, 10 pt, conference]{ieeeconf}  

\IEEEoverridecommandlockouts                              

\usepackage{graphicx}
\usepackage{amsmath} 
\usepackage{subcaption}
\usepackage{bm}
\usepackage{xcolor}

\usepackage{hyperref}
\usepackage{cleveref}
\usepackage{siunitx}
\Crefformat{figure}{#2Fig.~#1#3}
\Crefmultiformat{figure}{Figs.~#2#1#3}{ and~#2#1#3}{, #2#1#3}{ and~#2#1#3}

\graphicspath{{figures/}}

\DeclareMathOperator{\atantwo}{atan2}
\newcommand{\argmin}{\mathop{\rm arg~min}\limits}

\title{\LARGE \bf
SoPrA: Fabrication \& Dynamical Modeling of a Scalable Soft Continuum Robotic Arm with Integrated Proprioceptive Sensing
}

\author{
Yasunori Toshimitsu$^{1,2}$, Ki Wan Wong$^{1}$, Thomas Buchner$^{1}$, and Robert Katzschmann$^{1}$
\thanks{$^{1}$ ETH Zurich, Switzerland}%
\thanks{$^{2}$ The University of Tokyo, Japan}%
\thanks{{\tt\small \{\href{mailto:ytoshimitsu@ethz.ch}{ytoshimitsu},\href{mailto:kiwong@ethz.ch}{kiwong},\href{mailto:tbuchner@ethz.ch}{tbuchner},\href{mailto:rkk@ethz.ch}{rkk}\}@ethz.ch}}
}

\begin{document}

\maketitle
\thispagestyle{empty}
\pagestyle{empty}

\begin{abstract}
Due to their inherent compliance, soft robots are more versatile than rigid linked robots when they interact with their environment, such as object manipulation or biomimetic motion, and are considered to be the key element in introducing robots to everyday environments. 
Although various soft robotic actuators exist, past research has focused primarily on designing and analyzing single components. 
Limited effort has been made to combine each component to create an overall capable, integrated soft robot. Ideally, the behavior of such a robot can be accurately modeled, and its motion within an environment uses its proprioception, without requiring external sensors. 
This work presents a design and modeling process for a Soft continuum Proprioceptive Arm (\emph{SoPrA}) actuated by pneumatics. 
The integrated design is suitable for an analytical model due to its internal capacitive flex sensor for proprioceptive measurements and its fiber-reinforced fluidic elastomer actuators. 
The proposed analytical dynamical model accounts for the inertial effects of the actuator's mass and the material properties, and predicts in real-time the soft robot's behavior. 
Our estimation method integrates the analytical model with proprioceptive sensors to calculate external forces, all without relying on an external motion capture system.
\emph{SoPrA} is validated in a series of experiments demonstrating the model's and sensor's accuracy in estimation.
\emph{SoPrA} will enable soft arm manipulation including force sensing while operating in obstructed environments that disallows exteroceptive measurements.
\end{abstract}

\section{Introduction}
Soft robots are gaining popularity as an adaptable and versatile type of robot that can take advantage of its compliant body by succeeding in tasks that would be difficult for traditional rigid link robots \cite{rus_design_2015, schmitt_soft_2018}.
Soft continuum manipulators are particularly interesting, as they have similar functions as a traditional rigid link robot, but their inherent compliance ensures safe operation among humans and robust interaction with the environment.
They have been extensively explored in terms of actuator design \cite{marchese_recipe_2015, mosadegh_pneumatic_2014}, modeling and control schemes \cite{connolly_automatic_2016, katzschmann_dynamic_2019, katzschmann_dynamically_2019, du_diffpd_2021}, and sensorizations \cite{navarro_model-based_2020, truby_distributed_2020}.

However, when considering a soft robotic arm that can be used as a platform for application-oriented projects, all of these aspects must be combined into a unified design and model that can explain the deformation behavior of soft robots under internal actuation and external forces. We believe that truly proprioceptive soft robots are necessary to achieve improved manipulation and locomotion in order to someday introduce soft continuum robots into an everyday environment.

\begin{figure}[t]
    \centering
    \includegraphics[width=\columnwidth]{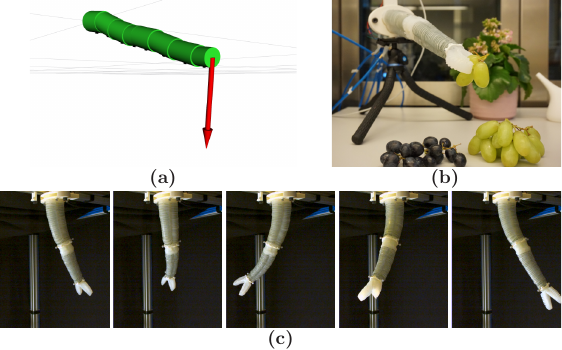}
    \caption{The design of the Soft continuum Proprioceptive Arm (\emph{SoPrA}) allows for (a,b) proprioceptive estimation of external forces, and (c) an analytical model to accurately predict its dynamic movement.}
    \label{fig:intro_figure}
\end{figure}

In this work, we introduce the design and modeling of a Soft continuum Proprioceptive Arm called \emph{SoPrA}. The creation of \emph{SoPrA}, which is shown in \Cref{fig:intro_figure}, is guided by three overarching principles:

\emph{Geometric Scalability:} The design should be easy to adapt to different sizes and numbers of segments to create bigger and longer arms with more actuated degrees of freedom (DOF). The main challenge with regard to scalability is maintaining the same fabrication methods while varying the pneumatic supply pathways between different segments and varying the parameters for different cross-sectional shapes.

\emph{Analytic Modelability:} The characteristics of the arm should be describable with an analytical, dynamical model that closely matches the actual behavior of the arm. The advantage is evident for proprioceptive model-based control, but the analytical model can also be used for data-driven models and controllers; they can be initialized with accurate model data, speeding up their learning process by focusing on higher-level tasks.

\emph{Full Proprioception:} The arm should have internal sensors that can take proprioceptive measurements of the arm pose. Even when external motion capture systems are not available, these sensors can enable shape reconstruction and external contact state estimation, which are necessary for a closed-loop control of a soft robot.

Pneumatic actuators have often been used for soft robotic actuator design due to their controllability and ease of fabrication, at the expense of requiring a separate pressure supply system \cite{marchese_recipe_2015}. 
One challenge in the design of pneumatic actuators for soft robots is how to model the deformation in the desired direction under pressurization. Some examples of designs for pneumatic actuators that have been extensively researched are pleated designs \cite{mosadegh_pneumatic_2014, katzschmann_autonomous_2015}, ribbed designs \cite{katzschmann_exploration_2018}, cylindrical designs \cite{marchese_autonomous_2014}, and fiber and ring-reinforced designs \cite{polygerinos_modeling_2015}.

Modeling of soft continuum arm robots can generally be subdivided into data-based and analytical model-based methods.
Data-driven models attempt to construct the model just from measurements from the actual robot \cite{qiao_dynamic_2019, bruder_modeling_2019}. However, extensive data gathering is required to capture the entire dynamics of the robot. 
There have been works which attempt to approximate the dynamic behavior by matching an augmented rigid body model \cite{katzschmann_dynamic_2019, della_santina_dynamic_2018}, or using Lagrangian dynamics \cite{falkenhahn_model-based_2015}.
Modeling of the deformation for a fiber-reinforced finger has been described in \cite{polygerinos_modeling_2015}.
Other approaches to modeling include linearization of FEM-based models \cite{katzschmann_dynamically_2019}, Cosserat rod models \cite{till_real-time_2019} or polynomial curvature fitting \cite{santina_control_2020}.

In soft robotics, different types of sensors have been exploited, but the integration of sensors into soft robotic manipulators has been limited. At one end of the cost spectrum, low-cost resistive flex sensors\,\cite{gerboni_feedback_2017, homberg_robust_2019} or stretch sensors\,\cite{nguyen_design_2017} are used. However, these type of sensors tend to have nonlinear characteristics with hysteresis, which makes the interpretation of these sensors problematic.
Kirigami-style cutouts of conductive silicone have a similar problem, the noisy sensor measurements required extensive data processing and a machine learning approach to decode and analyze the data\,\cite{truby_distributed_2020}.
At the other end of the spectrum are fiber-optic sensors, which can measure the continuous pose of the sensors\,\cite{sefati_fbg-based_2019}.
However, there are limited options for off-the-shelf fiber-optic shape sensors, and those are still prohibitively expensive.
Navarro et al. performed sensor fusion to estimate forces and deformations. Capacitive sensors find the contact location, and pneumatic sensors estimate the force intensity and the deformation, all by using an FEM-based numerical approach \cite{navarro_model-based_2020}.
Nguyen et al. has introduced conductive rubber stretch sensors into a single segment of a fiber-reinforced soft actuator, to measure the bending pose of the soft arm \cite{nguyen_design_2017}.

The contributions of our work are (a) a scalable design for a pneumatically actuated 3D fiber-reinforced soft continuum arm with integrated proprioceptive flex sensors; (b) an analytical model of a soft arm that considers the dynamic mechanical properties and the elastic properties of the soft silicone elastomer material; (c) a method to use the proprioceptive measurements from the flex sensors in the arm to estimate the external contact forces; and (d) extensive experimental evaluations of \emph{SoPrA}'s capabilities in terms of demonstrating modeled continuous motion and proprioception.



\section{Design and Fabrication Process}
\label{sec:design_fab}
\begin{figure}
        \centering
        \includegraphics[width=.9\columnwidth]{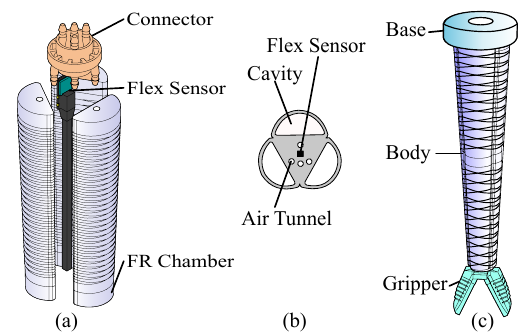}    
        \caption{Overview of the assembly process. (a, b) Three fiber-reinforced (FR) chambers are combined with a flex sensor to form a single actuated segment. The center of a segment also contains air tunnels, which are the pneumatic lines that connect the base to the distal chambers. 3D-printed connector pieces route the air tunnels between segments. (c) The segments are then combined with a soft pleated gripper and a base piece to form the completed arm.}
        \label{fig:fabrication_overview}
\end{figure}

\emph{SoPrA's} design is a soft robotic arm based on fiber-reinforced soft pneumatic actuators and a pleated soft pneumatic gripper without reinforcements. We produce the complete arm by fabricating individual fiber-reinforced (FR) chambers, combining three chambers into one segment, and combining the segments with proprioceptive sensors, pneumatic connectors, and a gripper. \Cref{fig:fabrication_overview} provides an overview of the assembly process. 

The cross section of each segment is tapered linearly from its base to tip. This design choice has several advantages in terms of fabrication and actuation. The taper acts as a draft feature, which eases the extraction of the chamber's silicone body from the 3D printed mold that forms the chamber's cavity. The center's triangular core contains the flex sensors and air tunnels that connect to the chambers in further distal segments. Since the more distal segments require fewer air tubes than the proximal segments, they can be slimmer.

The tapered design is also advantageous for the arm's mobility, since it allocates more actuation torque to the segments closer to the base. The base segments require higher actuation in order to lift the further distal segments. As there is an upper limit to the actuation pressure (beyond which the used material causes permanent deformation), a larger pressurized area must generate more bending torque. However, this comes at the expense of increasing the overall volume and, therefore, the weight of the actuator. A tapered design allows for the distal segments, which do not need to provide as much torque, to be narrower and lighter.

We cast each chamber separately, and individually wrap each one with fibers for radial reinforcement.
Three of the fiber-reinforced chambers are bundled together with a flex sensor and rigid rods that form the cavity for air tunnels, and silicone is poured into the middle to combine them into a single segment.

The flex sensors must not be stretched lengthwise because it could result in fatigue or inaccurate measurements. Thus, a mechanism that will prevent the sensor from being stretched lengthwise is included in the design. This mechanism can be taken into account in the analytical model; we can assume that the length of the soft arm's center axis remains constant.
Before a flex sensor is placed in the center of a segment without being stretched, it is first wrapped in low tack tape to protect it from the silicone, then in a piece of inextensible mesh fabric as is shown in \Cref{fig:bendlabs_cutout}. During the silicone curing, the silicone will seep through the mesh and become firmly bonded together with the mesh fabric. After the sensors have been embedded in the silicone, they can be re-calibrated to ensure an accurate reading.
The flex sensors use the I2C protocol to communicate, and four lines (VCC, ground, SDA, and SCL) run along the whole length of the arm to provide power and communicate with the sensors.

After each actuated segment has been fabricated, they are connected together using 3D-printed connectors, which route air tunnels between consecutive segments (as is shown in \Cref{fig:fabrication_overview}a). A stereolithography (SLA) 3D printer enables an intricate design, with empty tubular cavities that allow for air to pass through.  A soft pneumatic gripper is attached to the tip of the arm to enable object manipulation. For the gripper manufacturing process, we referred to the procedure from \cite{hao2016universal}. We manufactured the pleated fingers separately, which were then combined to form the gripper. A circular piece is attached to the base, which can be used as a flange to attach \emph{SoPrA} to fixtures.

\section{Analytical Modeling of SoPrA}
\label{sec:model}
Our analytical, dynamic model enables the proprioception of \emph{SoPrA} based on the measurements of its pressurization and bending.
The arm's shape is described using piecewise constant curvature (PCC) elements as approximation. Each actuated segment of the total $N_{seg}$ segments is equally split into a number of PCC elements, $N_{pcc}$. In PCC-based modeling of soft continuum arms, a single PCC element is usually assigned to a single actuated segment\,\cite{katzschmann_dynamic_2019}. However, when there are external forces such as gravity, this approximation does not hold, and the soft arm exhibits a varying curvature along a single segment.

In this work, we have increased the number of PCC elements in order to improve the accuracy of the model while retaining the simplicity of the PCC approximation. By balancing the introduced degrees of freedom, i.e., $N_{pcc}$, with the level of model complexity, i.e., PCC vs. Cosserat rod, we achieve sufficient accuracy while remaining performant when describing the robot's behavior.

\subsection{Overview of the model and parameterization of the shape}
The dynamic model for the soft arm is
\begin{equation}
        \label{eq:dynamic_model}
        A \bm{p} + J^T \bm f = B(\bm{q})\ddot{\bm{q}} + \bm c (\bm q, \dot{\bm q}) + \bm g (\bm q) + K \bm q + D \dot{\bm q}
\end{equation}
in which the matrix $A$ converts the chamber pressures to generalized forces, $J$ is the Jacobian, and $\bm f$ is the external force at the tip. $\bm p$ is a $3N_{seg}$-dimensional vector that represents the air pressure in each chamber, and $\bm q$ is the $2N_{seg}N_{pcc}$-DOF parametrization of the robot's pose. 
$B$, $\bm c$, and $\bm g$ are the inertia matrix, Coriolis \& centripetal force, and gravity, respectively. These are calculated from a matching rigid body model, which is explained in \Cref{sec:ara}.
$K$ and $D$ are the stiffness and damping matrices, respectively, and they are calculated from an analytical model of the bending moment, which is explained in \Cref{sec:silicone_model}.

The pose of a PCC element is usually described by the two variables $\phi$ and $\theta$, which are respectively the angle of the plane of bending and the curvature, as is shown in \Cref{fig:pcc_transform}. In this work, we introduce the variables $\theta_x$ and $\theta_y$ defined by the following equation:
\begin{equation}
        \begin{split}
                \theta_x &:= \theta\cos(\phi)\\
                \theta_y &:= \theta\sin(\phi)\\
        \end{split}
\end{equation}
This has the advantage that the parametrization does not reach singularity when the PCC section is straight. A similar approach to a singularity-free parametrization has been investigated in \,\cite{della_santina_improved_2020}, where the parameters were defined using arc lengths on the surface of a cylindrical PCC element.
It can be converted back to the $\phi$, $\theta$ coordinates through these equations:
\begin{equation}
        \begin{split}
                \phi &= \atantwo(\theta_y, \theta_x) \\
                \theta &= \sqrt{\theta_x^2 + \theta_y^2}
        \end{split}
\end{equation}

The pose of the entire soft trunk $\bm q$ is described by the longitudinal variables for each PCC element:
\begin{equation}
        \bm q = \begin{bmatrix} \theta_{x,1} && \theta_{y,1}  && \theta_{x,2} && \dots && \theta_{y,N_{seg}N_{pcc}} \end{bmatrix}^T
\end{equation}

\subsection{Kinetics through the Augmented Rigid Body Model}
\label{sec:ara}
\begin{figure}[htb]
\begin{subfigure}{0.45\columnwidth}
    \centering
    \includegraphics[width=.6\linewidth]{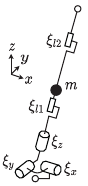}
    \caption{}
    \label{fig:rigid_link_model}
\end{subfigure}
\begin{subfigure}{0.45\columnwidth}
    \centering
    \includegraphics[width=\linewidth]{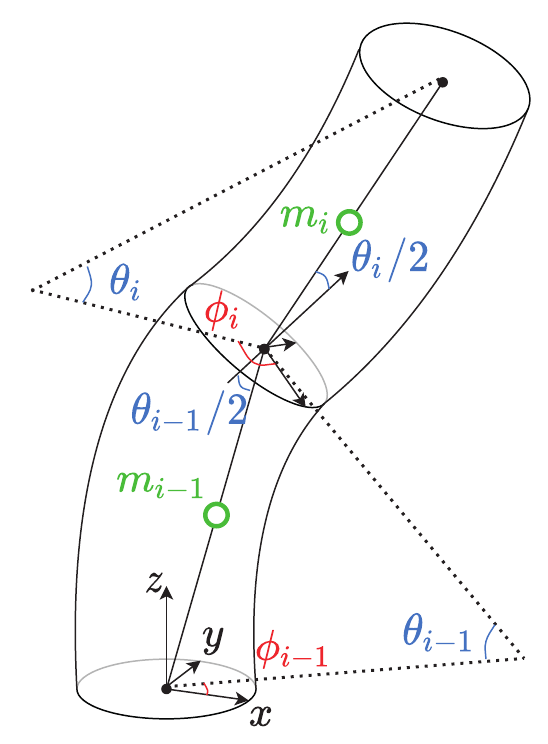}
    \caption{}
    \label{fig:pcc_transform}
\end{subfigure}
\caption{Kinematic representation. (a) A rigid link model showing the dynamics for a single PCC element. (b) Frame transformations between the bases of two PCC elements.}
\end{figure}

In order to calculate the kinetic elements $B$, $\bm c$, and $\bm g$ of \Cref{eq:dynamic_model}, an augmented rigid body model based on the one proposed in \cite{katzschmann_dynamic_2019} was used. Each PCC element has a rigid model as counterpart (see \Cref{fig:rigid_link_model}).
While the three-dimensional augmented rigid link model introduced in\,\cite{katzschmann_dynamically_2019} has $10$ joints per PCC section, the model in \Cref{fig:rigid_link_model} has been simplified to only contain $5$ joints. When fewer joints are used, rigid body algorithms become computationally faster.

Next, we will describe how the correspondence between the PCC pose $\bm q$ and the rigid link pose $\bm \xi$ was calculated. \Cref{fig:pcc_transform} describes $2$ PCC elements and the frame transformations between them. The rotational joints of the rigid link model for section $i$ must match the rotation between sections $i-1$ and $i$. The connector pieces and the end effector were modeled as PCC elements that are always kept straight, i.e., $\theta_x = \theta_y = 0$.
\begin{equation}
\label{eq:frame_calculation}
    R'\left(\phi_{i-1}, \frac{\theta_{i-1}}{2}\right) R'\left(\phi_i,\frac{\theta_i}{2}\right) = R_x(\xi_x) R_y(\xi_y) R_z(\xi_z)
\end{equation}
in which $R'(\phi, \theta)$ is the rotation $\theta$ around the vector $[-sin(\phi), cos(\phi), 0]^T$, and $R_x(\theta)$ is the rotation $\theta$ around the $x$ axis. The matrix equation in \eqref{eq:frame_calculation} was solved with a symbolic equation solver, to derive the relationship between the PCC pose $(\phi, \theta)$ and revolute joint poses of the rigid model $(\xi_x, \xi_y,\xi_z)$.
The prismatic joints $\xi_{l1}$ and $\xi_{l2}$ can be calculated as 
\begin{equation}
    \xi_{l1} = \xi_{l2} = \frac{l}{2} - \frac{l sin(\theta_i/2)}{\theta_i}
\end{equation}
in which $l$ is the length of the PCC element.

Using the augmented rigid body model, the kinetic parameters $B(\bm q)$, $\bm c(\bm q, \dot{\bm q})$, and $\bm g(\bm q)$ are calculated as in \cite{katzschmann_dynamic_2019}:
\begin{equation}
        \begin{cases}
                B(\bm q) &= J_m^T(\bm q) B_\xi(\bm m(\bm q)) J_m(\bm q)
                \\
                \bm c(\bm q, \dot{\bm q}) &= J_m^T(\bm q) \bm c_\xi(\bm m(\bm q), J_m(\bm q) \dot{\bm q})
                \\
                \bm g(\bm q) &= J_m^T(\bm q) \bm g_\xi(\bm m(\bm q))
        \end{cases}
\end{equation}
in which $\bm \xi = \bm m(\bm q)$ maps the configuration of the soft arm $\bm q$ to the configuration of the augmented rigid body model $\bm \xi$. $J_m$ is the Jacobian of $\bm m(\bm q)$. $B_\xi$, $\bm c_\xi$, and $\bm g_\xi$ are the kinetic parameters of the rigid model.

\subsection{Modeling of Stiffness and Damping}
\label{sec:silicone_model}
The stiffness and damping matrices of the segment, $K$ and $D$, can now be calculated from an analytical model of the soft arm's cross section.
These diagonal matrices designate the stiffness and damping of each PCC element.
The stiffness matrix is derived by considering the stress from the elastomer, which stretches and contracts when a PCC element is bent.

The fiber reinforcement around each chamber limits the radial expansion of the chambers, but it still allows for elongation in the longitudinal direction. The model assumes that the cross section shape is constant, regardless of the pressurization.
According to the derivation in\,\cite{polygerinos_modeling_2015}, our model considers only stress in the axial direction $s_1$. When a piece of elastomer stretches in the axial direction $\lambda$, the stress $s_1$ can be modeled as
\begin{equation}
\label{eq:stress_stretch}
    s_1 = \bar \mu  (\lambda - \frac{1}{\lambda ^3})
\end{equation}
in which $\bar \mu$ is the initial shear modulus. In \cite{connolly_automatic_2016}, this value was measured to be \SI{0.085}{\mega\pascal} for a silicone elastomer with Shore hardness 10A\footnote{Dragon Skin 10, Smooth-On Inc.}, which is the same material that has been used in this work. To account for any errors introduced by the approximations, we use for our model values of $\bar \mu$ that are characterized by the method described in \Cref{sec:characterization}.

\begin{figure}[htb]
    \centering
    \includegraphics[width=\columnwidth]{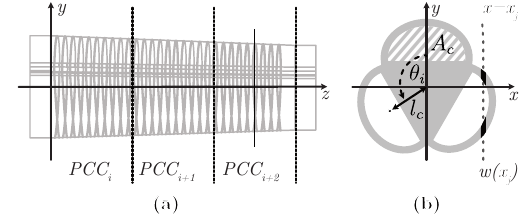}
    \caption{Subsections and cross section. (a) An arm's segment is subsectioned along the $z$-axis into three piecewise constant curvature (PCC) elements. (b) The cross section of one PCC element. Here the $y$ axis is the neutral axis of bending. It is spanned by the total width $w(x_j)$ at height $x_j$. $A_c$ is the cross-sectional area  of  the  chamber, $l_c$ is the distance between the centroid of the chamber to the center axis, and $\theta_j$ is  the  angle  between  the  $x$  axis  and  each  chamber.}
    \label{fig:cross_section_analytical}
\end{figure}

In \Cref{fig:cross_section_analytical}b, the cross section of a single PCC element within an actuated segment is shown.
Although each section is actually tapered, for simplicity each PCC element is modeled to have a constant cross section, which is represented by the cross section at the middle of the PCC element.

Since the center axis of the segment is constrained by an inextensible fabric, the plane that passes through the center axis is set as the neutral bending plane. The longitudinal stretch $\lambda(x)$ and strain $s(x)$ at position $x$ can be calculated with a linear approximation around $\lambda = 1$ by
\begin{equation}
        \begin{split}
                \lambda(x) &=  1 + \frac{\theta x}{l}
                \\
                s(x)&= \mu \left(\lambda(x) - \frac{1}{\lambda(x)^3}\right)
                \\
                &\approx 4\mu (\lambda(x) - 1) = 4\frac{\mu  x}{l}\theta
        \end{split}
\end{equation}
in which $l$ is the section length, and $\theta$ is the bending angle of the PCC section.
Therefore, the total moment $M_e$ generated by the elasticity of the silicone elastomer material is
\begin{equation}
\label{eq:elastic_moment}
\begin{split}
        M_e &= \int{x s(x) w(x) dx} = 4 \frac{\mu \theta}{l} \int{x^2 w(x) dx}
        \\
        &= \frac{4\mu  I}{l} \theta
\end{split}
\end{equation}
in which $w(x)$ is the total width spanned by the cross section at height $x$ (see \Cref{fig:cross_section_analytical}), and $I$ is the cross section's second moment of area.

The second moment of area $I$ is calculated at the center of each PCC segment. We first split the area into different portions whose moment of area can be analytically calculated, and then add them together. As the cross section has three symmetrical axes, the second moment of area around any orthogonal axis in the cross-sectional plane around the area's centroid remains the same.
Similar to the modeling of the moment $M_e$, the moment $M_d$, caused by dissipative forces, was described by
\begin{equation}
\label{eq:drag_moment}
    M_d = \frac{\rho  I}{l}\dot{\theta}
\end{equation}
in this work, in which $\rho$ is an unknown coefficient.

Finally, the moment $M_a$ generated by air pressure is
\begin{equation}
        M_a = \sum_{j=0}^3{p_j A_c l_c \sin(\theta_j)}
\end{equation}
in which $p_j$ is the air pressure of each chamber, $A_c$ is the cross-sectional area of the chamber, $l_c$ is the distance between the centroid of the chamber to the center axis, and $\theta_j$ is the angle between the neutral axis and each chamber (see \Cref{fig:cross_section_analytical}).

Therefore, we can obtain all the parameters in \Cref{eq:dynamic_model} and fully describe the dynamic model.

\subsubsection{Characterization of the Material Properties}
\label{sec:characterization}
The coefficients for the elastic and drag forces were determined by performing a characterization experiment on each assembled segment. Each segment is actuated in a feedforward manner through a pre-defined pressurization profile, which swings it back and forth for a cycle of three seconds. The curvature is measured with a motion capture system that uses reflective markers attached to the proximal and distal ends of each segment. For this experiment, constant curvature is assumed for a whole segment. To minimize the effect of gravity by adjacent segments, each segment is characterized separately before the entire arm is assembled.

The matrix $\mathcal{M_I}$ consolidates the known coefficients in \Cref{eq:elastic_moment} and \Cref{eq:drag_moment} as
\begin{equation}
    \mathcal{M_I} := \begin{bmatrix}
    I_1/l_1 & 0 & \dots & 0 \\
    0 & I_2/l_2 & \dots & 0 \\
    \vdots & \vdots & \ddots & \vdots \\
    0 & 0 & \dots & I_N/l_N 
    \end{bmatrix}
\end{equation}
Then, the known forces can be combined into
\begin{equation}
    \bm f_\text{known} := A \bm p + J^T \bm f - B \ddot{\bm q} - \bm c - \bm g.
\end{equation}
The velocity $\dot{\bm q}$ and acceleration $\ddot{\bm q}$ are obtained from a 5th-order polynomial approximation of the position $\bm q$ of a single swing movement of the arm.
To best fit the $N_s$ observed data samples of $\mathcal{M_I}$, $q$, and $f_\text{known}$, the unknowns $\mu$ and $\rho$ are calculated by
\begin{equation}
    \begin{bmatrix}
    \mu \\ \rho
    \end{bmatrix}
    =
    \begin{bmatrix}
    4 \mathcal{M_I} \bm q_1 & \mathcal{M_I} \dot{\bm q}_1 \\
    \vdots & \vdots \\
    4 \mathcal{M_I} \bm q_{N_s} & \mathcal{M_I} \dot{\bm q}_{N_s}
    \end{bmatrix} ^ \dagger
    \begin{bmatrix}
    \bm f_{\text{known}, 1} \\
    \vdots \\
    \bm f_{\text{known}, N_s}
    \end{bmatrix}
\end{equation}
in which $^\dagger$ indicates the pseudoinverse.

\section{Proprioceptive Flex Sensor}
\label{sec:sensor}
\begin{figure}[h]
\centering
\begin{subfigure}{0.61\columnwidth}
    \includegraphics[width=\linewidth]{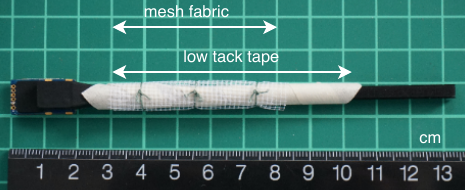}
    \caption{}
    \label{fig:bendlabs_cutout}
\end{subfigure}
\begin{subfigure}{0.29\columnwidth}
    \includegraphics[width=\linewidth]{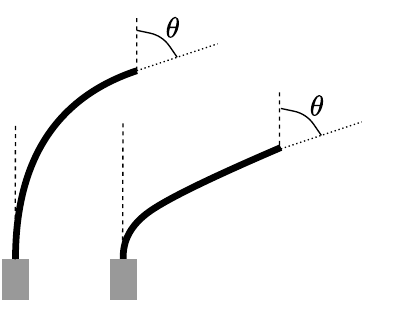}
    \caption{}
    \label{fig:BendLabs_sensormodel}
\end{subfigure}
\caption{Flex sensor. (a) A 2-axis digital flex sensor from Bend Labs. During the fabrication of the sensor, the entire sensor is covered with low tack tape and mesh fabric. Here it is partially covered for demonstration. (b) A sensor model of the flex sensor. Both sensor shapes output the same measurement because the measurement modality is independent of the sensor's bend shape; it only depends on the relative orientation of the beginning and end.}
\end{figure}

We describe the sensor model and propose a method to integrate the measurements from the proprioceptive data into the dynamic model so as to gain information about its contact state.
We use a 2-axis digital flex sensor\footnote{Bend Labs Inc., \url{https://www.bendlabs.com}} as shown in \Cref{fig:bendlabs_cutout} for proprioceptive measurements.
This sensor measures the bending angle between its two ends. The sensor's measurements are not affected by the sensor path, as it is demonstrated in \Cref{fig:BendLabs_sensormodel}.

Capacitive flex sensors are relatively inexpensive\footnote{US \$129 as of the day of manuscript submission}, and an exact sensor model is known.
For example, when a sensor spans across three PCC elements, the following relationship can be described:
\begin{equation}
    \begin{split}
        s_x &= \theta_{x,1} + \theta_{x,2} + \theta_{x,3}
        \\
        s_y &= \theta_{y,1} + \theta_{y,2} + \theta_{y,3}
    \end{split}
\end{equation}
in which $s_x$ and $s_y$ are the measurements of each axis of the sensor.
Expanding on this, the linear relationship between the soft robot's pose $\bm q$ and the sensor measurement $\bm s$ can be written as:
\begin{equation}
\label{eq:sensor_model}
    \bm s = S \bm q
\end{equation}
A single flex sensor was placed in each segment; therefore, $S$ is a $2N_{seg}\times 2N_{seg}N_{pcc}$ matrix. For $N_{pcc} > 1$ it is not invertible, so the exact pose of the robot cannot be determined from the sensor measurements alone. However, the measurement can be used as a linear constraint, which can be combined with the analytical dynamic model in \Cref{eq:dynamic_model} to gain further information about the robot's state.

\subsection{Contact force information from sensor measurements}
We propose a method to combine the measurements of the flex sensor with our dynamic model to determine the tip contact state of the soft robotic arm. For simplicity, we assume that the robot is static during the proprioceptive measurement, and no dynamical effects need to been considered, i.e., $\dot{\bm q} = \ddot{\bm q} = \bm c (\bm q, \dot{\bm q}) = \bm 0$.

\subsubsection{Model Error Improvement through Sensor}
We describe a method to improve the model error by using the sensor measurements.
This is done by solving a quadratic program, which tries to find a robot's pose $\bm q$ that can most accurately explain the sensor measurements:
\begin{equation}
\label{eq:optimize_q}
    \begin{split}
        \argmin_{\bm q}& \quad || A \bm p - \bm g(\bm q_0) - K \bm q || ^2
        \\
        \text{subject to}&\quad \bm s = S \bm q
    \end{split}
\end{equation}
in which $||\cdot||^2$ describes the L2 norm, and $\bm q_0$ is the seed pose that distributes the sensor measurements equally to each PCC element within that sensor, which can be described as
\begin{equation}
    \bm q_0 = S^\dagger \bm s
\end{equation}
The pose gained as a result of this optimization corrects the model error.

\subsubsection{External Force at Tip through Sensor}
We can also include the external force $\bm f_{ext}$ at the tip of the arm into our optimization:
\begin{equation}
\label{eq:optimize_q_f}
    \begin{split}
        \argmin_{\bm q, \bm f_{ext}} \quad (||& A \bm p - \bm g(\bm q_0) + J(\bm q_0)^T \bm f_{ext} \\
        & - K \bm q + \bm f_0 ||^2 + r ||\bm f_{ext} || ^2 )
        \\
        \text{subject to}\quad &\bm s = S \bm q
    \end{split}
\end{equation}
The generalized force $\bm f_0$ is an offset, which is measured before an external force is applied to the robot. This generalized force is defined by
\begin{equation}
\label{eq:f_0}
    \bm f_0 := -A \bm p + \bm g(\bm q)
\end{equation}
in which $\bm q$ is the robot pose obtained from \Cref{eq:optimize_q}. This correction in model and sensor error can be described by an offset force in the generalized coordinates; one can consider this to be the tare weight measurement.

\section{Experimental Results}
\label{sec:experiments}
We conducted the experiments on a two-segment configuration of \emph{SoPrA}. Each of the six chambers and the gripper were pneumatically actuated with a controlled proportional valve manifold\footnote{MBA-FB-VI, \SI{0}-\SI{2}{bar} range, 1\% accuracy, by Festo SE \& Co. KG}. The properties of the silicone elastomer were characterized via the process described in \Cref{sec:characterization}. $(\mu, \rho) = (\SI{43000}{\pascal}, \SI{61000}{\pascal\second})$ for the upper segment, and (\SI{57000}{\pascal}, \SI{8000}{\pascal\second}) for the lower segment. The robot motion library called \emph{Drake} was used for the calculation of the inertial parameters of the rigid body model\,\cite{drake}.

\subsection{Verification of the accuracy of the dynamic model}

\begin{figure}[h]
        \centering
        \includegraphics[width=0.9\columnwidth]{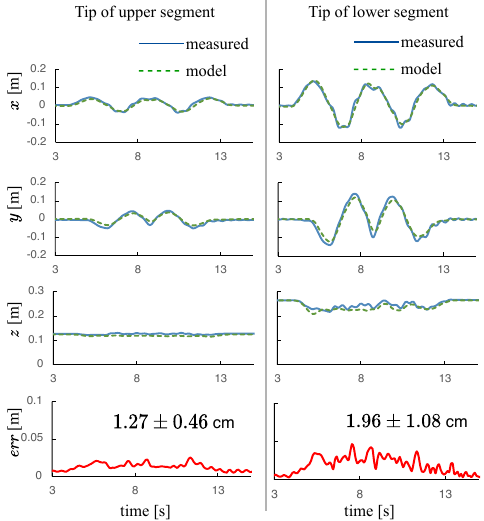}
            \caption{Results of the comparison between the tip position measured on the robot with a motion tracking system and the simulation with the dynamic model, for $N_{pcc} = 3$ PCC elements per actuated segment. The error $\textit{err}$ is defined in \Cref{eq:error}.}
        \label{fig:model_measurement_comparison_just_graph}
\end{figure}

\begin{figure}[htb]
    \centering
    \includegraphics[width=0.85\columnwidth]{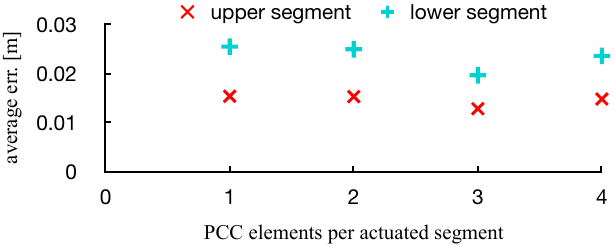}
    \caption{Change in model accuracy for different numbers of PCC elements per actuated segment. The error $\textit{err}$ is defined in \Cref{eq:error}, and the average is calculated for the duration of the movement.}
    \label{fig:N_PCC}
\end{figure}

In this experiment, \emph{SoPrA} was actuated with a feedforward pressure profile consisting of sinusoidal sweeps. This commanded pressure profile was fed into the dynamic model of \Cref{eq:dynamic_model} as $\bm p$, and $\ddot{\mathbf q}$ was integrated. Since an efficient soft robotic simulation system is beyond the scope of this work, the integration was achieved by implementing a simple Euler-Richardson method with a time step of \SI{0.02}{\milli\second} to stabilize the simulation.
A reflective marker-based motion capture system\footnote{Miqus M3, Qualisys AB} was used to take ground truth measurements of the soft arm's movement.
The error between the ground truth position of each segment's tip position and the position calculated from the dynamic model \Cref{eq:dynamic_model} gives a quantitative measure of the accuracy of the model: 
\begin{equation}
\label{eq:error}
    err := | \bm p_{meas}- \bm p_{model} |
\end{equation}
To verify that the model can capture the dynamic properties of \emph{SoPrA}, it was actuated to move at a cycle time of \SI{3}{\second} with a delta in lower segment tip position of about \SI{30}{\centi\meter}. The entire actuation sequence of the experiment, overlayed with the data from the model, can be seen in the video attachment.

The results in \Cref{fig:model_measurement_comparison_just_graph}show the scenario when three PCC elements were assigned to each actuated segment (i.e., when  $N_{pcc} = 3$). The position error between the measurements and model was, on average, \SI{1.27}{\centi\meter} for the tip of the upper segment, and \SI{1.96}{\centi\meter} for the tip of the lower segment. 

To determine the optimal $N_{PCC}$ (i.e., the number of PCC elements per actuated segment), we have run the same forward simulation with the same pressure profile as the one in \Cref{fig:model_measurement_comparison_just_graph} for different $N_{PCC}$. The results are shown in \Cref{fig:N_PCC}.
The average error in the estimation of the tip position decreased as $N_{PCC}$ was increased up to $N_{PCC} = 3$, but the error rose again at $N_{PCC} = 4$.


\subsection{Estimation of external load using proprioceptive measurements}

\begin{figure}[htb]
    \centering
    \includegraphics[width=0.95\columnwidth]{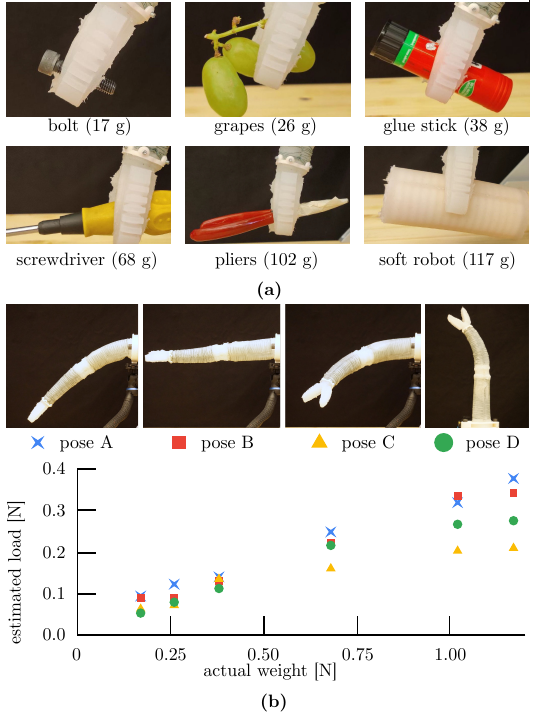}
    \caption{External load estimation experiment. (a) Objects that were grasped by the soft arm. (b) Arm poses in which the load was measured, and their estimation result.}
    \label{fig:load_experiment}
\end{figure}

In this experiment, \emph{SoPrA} was subject to an external force by picking with its gripper various objects of different weights in the range between \SI{17}{\gram} and \SI{117}{\gram}.
After the object is placed in the robot's gripper and a static equilibrium is reached, the quadratic program described in \Cref{eq:optimize_q_f} was used to estimate the load at the tip of the soft arm.
The grasped objects are shown in \Cref{fig:load_experiment}a. \emph{SoPrA} was able to grasp all objects without any physical modifications. To evaluate the estimation performance in various configurations, each measurement was taken in four different poses, all of which are shown in the upper part of \Cref{fig:load_experiment}b.
The results are shown in lower part of \Cref{fig:load_experiment}b, where the actual weight of the object is plotted against the estimated weight, defined by $|\bm f_{ext}|$, in the four poses.

\section{Discussion}
The characterized shear modulus $\mu$ is of the same order as the one introduced in \cite{connolly_automatic_2016} (\SI{85000}{\pascal}). The much higher coefficient $\rho$ of the dissipative moment $M_d$  observed for the upper segment ($7.6$ times larger than the lower segment) could be explained by the larger chamber cavity, which takes longer to pressurize and will show up as a dissipative effect under the current formulation. In order to precisely model fast movements, a model of the pneumatic supply may be required.

The model was able to predict the tip position during a feedforward actuation movement with a cycle of \num{3} seconds, at an average accuracy of \SI{1.96}{cm}. Considering that the tip-to-tip distance of the fingers of the open gripper is \SI{4}{cm}, we believe that this model accuracy is enough to conduct simple gripping tasks with feedforward control. The prediction accuracy has improved slightly by increasing $N_{PCC}$, however there was only a \SI{22.8}{\percent} improvement in accuracy when changing $N_{PCC}=1$ to $N_{PCC}=3$. An increase in $N_{PCC}$ computationally costs approximately $O(N_{PCC}^2)$, whereas much of the overhead has been identified to come from the explicit computation of the inertia matrix in the software Drake\,\cite{drake}, which uses the composite body algorithm\,\cite{featherstone2014rigid}. In future work, an adaptive model could minimize the computational complexity by adjusting in real-time  $N_{PCC}$ to be just sufficient to achieve the desired task.

In the load estimation experiment, \emph{SoPrA} attempted to estimate the external load on the tip. The robot did not accurately estimate the absolute weight of each object. For example, the \SI{102}{\gram} pliers would cause a \SI{1}{\newton} load, but only around \SI{0.3}{\newton} of force was estimated. 
This may be due to the elastic parameters being identified while the robot actuates itself (Section \ref{sec:characterization}), not under a known external load, so it may be insuitable for estimating external loads.
However, there was a linear relationship between the actual weight and the estimated load, and therefore the robot was able to measure the relative weight of each object. The estimated load was relatively consistent for all the poses used in the measurement.

\section{Conclusion}
We have presented the design and modeling of \emph{SoPrA}, a pneumatically actuated soft continuum arm robot with integrated proprioceptive sensors. The use of fiber reinforcement restricts the expansion of the pneumatic chambers in a radial direction, and this means that an analytical model can be used to describe the bending behavior of the robot. Furthermore, we have introduced a concept that describes the kinematics and dynamics of a PCC element through a simplified configuration, using a reduced number of joints in the augmented rigid body model\,\cite{della_santina_model-based_2020,katzschmann_dynamic_2019}.
Together with these enhancements, we can arbitrarily increase the number of PCC elements assigned to a single actuated segment. This enhancement can describe the configuration of the soft arm in a more precise manner when a single segment bends with a non-constant curvature. With these changes, we improved the accuracy of the model by about \SI{22.8}{\percent} in terms of the tip position error between the actual robot's movement and the simulation.

Moreover, we can take proprioceptive shape measurements of the robot's pose by integrating a capacitive flex sensor into the center axis of the arm during the assembly process. There is a linear relationship between the sensor's measurements and the arm's pose, and we have introduced a method that can combine the sensor's measurements with the model to correct for the model error and estimate external contact states.

We believe that, with this modelable, scalable, and proprioceptive design, soft continuum manipulators can break free from external sensing systems; utilizing its dynamic model, more task-oriented research on soft robots will be possible.
In the future, we will continue to improve the design and model formulation of \emph{SoPrA}, potentially introducing more modes of proprioception and increasing the accuracy of the model. We will also explore the application of \emph{SoPrA} to practical tasks in everyday environments such as households or workplaces and investigate which further advancements in design, modeling, and control are required for soft robots to assist humans better.



\bibliography{IEEEabrv,IEEEexample,references,SRL-ETH}
\bibliographystyle{IEEEtran}

\end{document}